\documentclass[Afour,sagej,times]{sagej}

\usepackage{moreverb,url}

\usepackage[colorlinks,bookmarksopen,bookmarksnumbered,citecolor=red,urlcolor=red]{hyperref}
\usepackage{subfig}
\usepackage{tikz}

\graphicspath{{figures/}{pictures/}{images/}{./}}

\newcommand\BibTeX{{\rmfamily B\kern-.05em \textsc{i\kern-.025em b}\kern-.08em
T\kern-.1667em\lower.7ex\hbox{E}\kern-.125emX}}

\begin{document}

\runninghead{de Leeuw den Bouter et al.}

\title{ProtoExplorer: Interpretable Forensic Analysis of Deepfake Videos using Prototype Exploration and Refinement}

\author{Merel de Leeuw den Bouter\affilnum{1,2}, Javier Lloret Pardo\affilnum{1}, Zeno Geradts\affilnum{1,2} and Marcel Worring\affilnum{1}}

\affiliation{\affilnum{1}Informatics Institute, University of Amsterdam, The Netherlands\\
\affilnum{2}Netherlands Forensic Institute, The Netherlands}

\corrauth{Merel de Leeuw den Bouter, Informatics Institute, University of Amsterdam, Science Park 900, 1098 XH Amsterdam, The Netherlands.}

\email{m.l.deleeuwdenbouter@uva.nl}

\newcommand{\mw}[1]{\todo[inline,color=magenta!40,author=Marcel]{ #1}}
\definecolor{actions_color}{RGB}{0, 179, 210}
\definecolor{indirect_actions_color}{RGB}{122, 185, 42}
\definecolor{findings_color}{RGB}{233, 74, 82}
\definecolor{exploration_loop_color}{RGB}{92, 35, 14}

\definecolor{candidate_to_be_discarded}{RGB}{196, 196, 196}

\definecolor{pristine_color}{RGB}{33, 161, 220}
\definecolor{manipulated_color}{RGB}{228, 16, 37}

\begin{abstract}
In high-stakes settings, Machine Learning models that can provide predictions that are interpretable for humans are crucial. This is even more true with the advent of complex deep learning based models with a huge number of tunable parameters. Recently, prototype-based methods have emerged as a promising approach to make deep learning interpretable. We particularly focus on the analysis of deepfake videos in a forensics context. Although prototype-based methods have been introduced for the detection of deepfake videos, their use in real-world scenarios still presents major challenges, in that prototypes tend to be overly similar and interpretability varies between prototypes. This paper proposes a Visual Analytics process model for prototype learning, and, based on this, presents ProtoExplorer, a Visual Analytics system for the exploration and refinement of prototype-based deepfake detection models. ProtoExplorer offers tools for visualizing and temporally filtering prototype-based predictions when working with video data. It disentangles the complexity of working with spatio-temporal prototypes, facilitating their visualization. It further enables the refinement of models by interactively deleting and replacing prototypes with the aim to achieve more interpretable and less biased predictions while preserving detection accuracy.
The system was designed with forensic experts and evaluated in a number of rounds based on both open-ended think aloud evaluation and interviews. These sessions have confirmed the strength of our prototype based exploration of deepfake videos while they provided the feedback needed to continuously improve the system. 
\end{abstract}


\maketitle

\section{Introduction}
Due to the growing fidelity of deepfake videos, it is becoming increasingly difficult to determine whether a video has been manipulated. As such, they pose a serious risk to our society. Tools for \textit{deepfake generation} are more accessible than ever, including open source tools \cite{DBLP:journals/corr/abs-2005-05535}, and easy-to-use apps \cite{Reface.Videos, AvatarifyLife} that do not require expert knowledge. In this paper we focus on video forensics, where the aim is to make judgments on the authenticity of a video which can be defended in court. The high fidelity of deepfakes, combined with their proliferation and the accessibility of the tools to generate them, presents a challenge as investigators are expected to get confronted with large volumes of disputed material.  

In recent years, in parallel with methods for deepfake generation, we have witnessed a rise in the availability of \textit{deepfake detection} methods. Open competitions in Kaggle such as the Deepfake Detection Challenge (DFDC), associated with the release of the dataset of the same name \cite{Dolhansky2020TheDD}, have contributed to the availability of deepfake detectors that were open sourced as an outcome of the competition. Although the release of these detectors may initially seem to address the needs of video forensic experts, the rules of the competition led to completely automated detectors that do not allow for input from the expert. 

In forensics, the video expert, in search of visual evidence that can be presented in court, conducts an in-depth analysis of the parts of the video that are in dispute. In this process, the expert looks for subtle indicators such as face wobble, blurred edges, missing facial features, or overly smooth skin \cite{Westerlund_review}.  Modern automatic deepfake detectors are more and more based on Deep Neural Networks (DNNs) as they outperform traditional methods  used in video forensics \cite{Rossler2019FaceForensics++,Verdoliva2020MediaOverview}. The opaque nature of DNNs is a cause of concern when used in a high-stakes setting such as forensics. In general DNNs do not allow for interaction with the internals of the model and even if that is the case, many of them are not interpretable in terms of the indicators forensic experts use. 

This concern regarding explainability has led to the development of methods that aim to provide more understandable predictions \cite{Selvaraju2017Grad-CAM:Localization, Binder2016Layer-wiseLayers, DeepLIFT, IntegratedGradients, LIME}.
An increasingly common approach is prototype learning \cite{Ras2020ExplainableUninitiated}. In image recognition with DNNs, \textbf{prototypes} are representative image patches of samples from the training set. These prototype-based methods are inspired by the ``case-based reasoning'' paradigm \cite{Agnar1994Case-BasedApproaches, Rudin2021InterpretableChallenges}, commonly used by humans to solve problems \cite{AllenNewell1972HumanSolving}. This paradigm tries to solve a current problem using solutions to similar previous problems. This approach also tends to be used by humans when recognizing objects. Studies in visual recognition \cite{Edelman1995, Edelman1992} indicate that humans recognize objects and patterns ``by matching the visual information with internally stored, viewpoint-specific exemplars, or prototypes'' \cite{Ware2004}. Network architectures that include prototypes, being closely related to the human reasoning process, can prove helpful in making the network more interpretable.

For video forensics, prototype-based learning methods have been developed \cite{Trinh2021InterpretablePrototypes} by building on methods that use prototypes for image classification \cite{Li2018DeepPredictions, Chen2018ThisRecognition} which provide predictions based on similarity to these prototypes. For video, the prototypes are \textbf{dynamic}, corresponding to a spatio-temporal volume in one of the video fragments used to train the model. This introduces a number of challenges. First, working with dynamic prototypes is more complex, because they have both a spatial and a temporal dimension. Secondly, manually performing prototype-based predictions for long video sequences, without specific tools for visualizing and temporally filtering them, is challenging. Lastly, the proper selection of prototypes is key, not only for the performance of the model but also for its explainability. In general prototype learning, prototypes are automatically identified by the network in the training process. This leads to a selection of prototypes that are suitable for representation learning but which are not ideal in terms of explainability. The input from the video forensic expert is therefore fundamental, and we argue that, for that purpose, a Visual Analytics (VA) system is needed. \textbf{Visual Analytics} should allow the video forensic expert to refine the selection of prototypes with the goal of improving or at least maintaining the performance of the model while maximizing the explainability of its predictions.

There are few existing VA systems for interactive prototype learning. The early method proposed in \cite{Migut2011InteractivePrototypes} for prototype exploration in relation to performance was conceived before DNNs became the prime image recognition model and is therefore less suitable for our purpose. In \cite{Zhao2020ProtoViewer:Prototypes}, a post-hoc approach is used \cite{ProtoFac}. Their method takes the embedding layer in a network and factorizes it into a number of prototypes. Thus, they build a surrogate model that mimics the behavior of the analyzed model. This has the advantage that it can be applied to any existing deep learned model. The prototypes, however, are not learned during the training of the original network. Hence, their approach does not lead to a visual analytics solution in which the user can interact with the model directly.  In \cite{Li2018DeepPredictions, Chen2018ThisRecognition}, prototypes were incorporated directly in the DNN architectures, to make interpretable predictions a characteristic of the model rather than a post-hoc explanation. But these have not been embedded in a visual analytics solution. None of these approaches are suitable when analyzing video data and using more complex dynamic prototypes that contain spatial and temporal information.

Based on the above considerations, we introduce ProtoExplorer, a VA system for prototype-based exploration and refinement of neural networks in the forensic analysis of deepfake videos. This system allows forensic experts to:

\begin{itemize}
\item \textbf{Conduct deepfake analysis on long videos} by using temporal filtering tools, specifically designed for prototype-based models to simplify model inference and interpretation of the predictions when working with video data.
\item \textbf{Visually explore prototype-based detection models}  by intuitively switching between multiple representations, each of which offers complementary information.
\item \textbf{Improve the explainability of deepfake detection models} by manually deleting prototypes from the model or by replacing prototypes based on suggested alternative prototypes and, after each refinement, calculating and visualizing the impact of the changes on the performance and characteristics of the model.
\end{itemize}

In summary, ProtoExplorer targets a specific task, i.e., the forensic analysis of deepfake videos, designed with and for video forensic experts. Our VA system enhances their expertise with meaningful visualizations and intuitive interactions with prototype-based models. Our contributions include:

\begin{itemize}
  \item A \textbf{VA model for interacting with prototype based networks} specifically designed for deepfake analysis but suitable for more generic image/video based tasks. 
  \item A \textbf{VA system for deepfake detection based on prototype learning}, designed to allow the video forensic expert to explore and refine the dynamic prototypes of the model with the goal of improving performance, as well as explainability and fairness.
  \item A thorough \textbf{evaluation of the system} with video forensic experts.
\end{itemize}

\section{Related Work}
Ever since the start of the deep learning era, the opaque nature of neural networks has been a concern and has been addressed both in the visual analytics community as well as in the machine learning community. Before discussing related VA systems for prototype refinement, we will introduce relevant existing methods in the realm of visual analytics systems for AI and interpretable and explainable AI.

\subsection{Visual analytics systems for AI}
Building interactive systems for understanding deep learning models has been an active research topic in the visual analytics community as the inherent complexity of deep neural networks and the overwhelming number of parameters makes it impossible for humans to easily understand the inner working of the network. The layered architecture of networks helps in this respect, as it provides natural levels of abstraction, but due to the the huge number of parameters it remains challenging. A number of excellent surveys on the topic have appeared, such as \cite{ALICIOGLU2022502} and \cite{Yuan_survey2021}. We are particularly interested in prototype-based methods. In the references, the only visual analytics system targeting prototype-based deep neural networks is ProtoSteer \cite{Ming_protosteer} which targets recurrent neural networks, a specific class of deep sequence networks. 

Apart from the concerns pertaining to the black-box nature of neural networks, there is an increasing awareness that the training data and the way the networks learn the model might lead to biased results. Typical observations, found when analyzing results of DNNs are biases towards specific gender, race, or demographics. Recently, visual analytics solutions have been proposed that aim to make this bias visible \cite{Munechika_bias,KWon_bias}. The problem of bias has also been identified for deepfake detection \cite{FairnessDeepfakeDatasets2021}. Our method aims to reduce the bias by letting experts refine the set of prototypes to assure that they are diverse and representative.

\subsection{Interpretable and Explainable Machine Learning}
In recent years, there has been increased attention for methods that help make predictions from Machine Learning (ML) models more understandable to humans. Yet, there is a lack of consensus in literature \cite{Rudin2021InterpretableChallenges, Ras2020ExplainableUninitiated, MolnarInterpretableLearning} regarding the differences between a ML model being interpretable or explainable. In this paper, we use the term interpretability based on the definition in \cite{MolnarInterpretableLearning, Miller2019} that defines interpretability as `the degree to which a human can consistently predict the model’s result'. 

Methods in this field can also be distinguished based on whether they are intrinsic or post-hoc. Post-hoc methods for DNNs can be applied to a model after it has been trained. There are various approaches which use post-hoc methods, with the most prominent being visualization methods based on back-propagation \cite{Selvaraju2017Grad-CAM:Localization, Binder2016Layer-wiseLayers, DeepLIFT, IntegratedGradients}. The fact that post-hoc methods can be applied to a model after training makes them very flexible and convenient. However, recent research has demonstrated that post-hoc visualization methods are not always reliable in their visual explanations \cite{Adebayo2018SanityMaps}. Another post-hoc approach is to build a surrogate interpretable model that locally approximates the original model. This surrogate model can explain the predictions of a classifier or regressor in a faithful way \cite{LIME}. This approach adds an extra layer of complexity, making it harder to troubleshoot as it can be unclear whether the opaque model or the surrogate model is wrong \cite{Rudin2021InterpretableChallenges}.

Intrinsic methods, when based on DNNs, modify the architecture of the network before training the model in order to provide more understandable predictions. Among intrinsic methods that have recently been introduced, prototype-based methods stand out in the image recognition field because of their potential to provide explainability based on visual similarity \cite{Chen2018ThisRecognition}. This approach based on case-based reasoning relates to the way humans recognize objects and subjects. Since then, prototype learning methods have been developed, including methods that use decision trees \cite{Nauta2021NeuralRecognition}, hierarchically structured prototypes based on predefined taxonomies \cite{Hase2019InterpretablePrototypes}, and deformable prototypes that adapt to pose variations of objects \cite{Donnelly2022}. 

Prototype-based methods are also used in high-stakes settings, in which their more interpretable predictions are very valuable. One example is the use of prototypes for medical diagnoses based on X-ray imagery \cite{Kim2021XProtoNet:Explanations}. In video forensics, and more specifically in deepfake detection tasks, the subject of analysis is not a still image but a video. DPNet \cite{Trinh2021InterpretablePrototypes} was conceived for the detection of deepfake videos. It uses dynamic prototypes containing not only spatial but also temporal information provided by pre-computed optical flow fields. DPNet is, to our knowledge, the only existing prototype-based neural network specifically conceived for deepfake detection. Our VA system provides a series of functionalities that allow the forensic expert to refine the prototype selection of models trained with DPNet and hence directly interact with the model.

\subsection{Refinement of Prototype-based Models}
As automatic prototype selection methods are not always optimal, different approaches have recently been introduced to refine the selection of prototypes. Chen et al. \cite{Chen2018ThisRecognition} introduce a post-training pruning method to remove “background” prototypes from the model. Ming et al. \cite{Ming2019InterpretablePrototypes} provide a task-agnostic method that allows the user to create, revise and delete prototypes of the model. These methods are either fully automated, without expert user input, or are task-agnostic, i.e. not designed for working with prototypes conceived for a specific task like deepfake detection. None of the current methods offers a VA system that allows a suitable workflow for model inference and prototype refinement.

Despite the proliferation of VA systems that aim to increase the explainability of DNNs in multiple tasks \cite{Spinner2019ExplAIner:Learning, ExplainExplore2020, Zhao2022Human-in-the-loopModels, Alperin2020ImprovingVisualizations, SCANViz2020, Summit2020, VATLD2021, Cheng2021VBridge:Models, Jin2022}, there is a lack of VA systems for prototype-based models. The only VA system for prototype refinement of neural networks that we are aware of \cite{Zhao2020ProtoViewer:Prototypes}, was not conceived to work with temporal media such as deepfake videos, nor with prototypes that combine spatial and temporal information. More importantly, it follows a post-hoc approach, building a surrogate model that mimics the behavior of the analyzed one \cite{ProtoFac}. This approach can only indirectly interact with the model as the prototypes are only part of the surrogate model. If by examining the prototypes and their contribution to predictions the user decides that the model has to be updated, they can only do so via the hyperparameters which have no direct connection to the prototypes. Our system uses prototype-based networks as backbone, and hence, it allows the user to directly analyze and refine prototype-based models without the need for surrogate models. Furthermore, our system targets a specific task, deepfake detection, and is specifically designed to manipulate the more complex prototypes used in this task. 

\section{Background}
\begin{figure*}[ht]
  \centering
  \includegraphics[width=\linewidth]{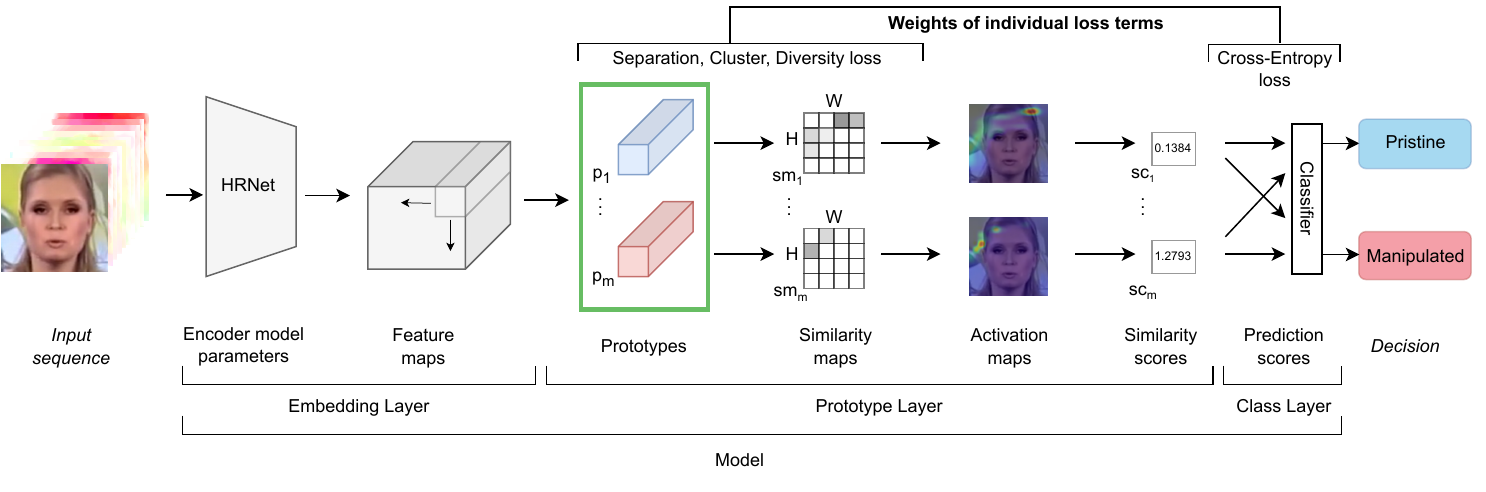}  \caption{Architecture of the deepfake detection model DPNet. Based on a large set of \textcolor{pristine_color}{pristine} and \textcolor{manipulated_color}{manipulated} video fragments, the feature encoder produces an embedding in the latent space. This space is used to select a set of prototypical training samples. The final classification of a train/test video is done in the class layer based on the similarities of the sample to the prototypes. Before training, the \textbf{weights corresponding to the individual loss terms} can be adjusted, to create balance between the terms. The prototypes (in the green box) are the basis for the interaction between the model and the human expert. A change in prototypes will cause a change in the loss function as well.}
  \label{fig:dpnet-arch}
\end{figure*}

The backbone of our VA solution is an automatic deepfake detection network, in particular the Dynamic Prototype Network (DPNet) \cite{Trinh2021InterpretablePrototypes}. Here we elaborate on its structure and parameters, illustrated in Figure~\ref{fig:dpnet-arch}, as these provide the levers which we can manipulate and which we need to visualize in our ProtoExplorer system. 

\subsection{Network architecture}
Deepfake detection methods take a video sequence as input and make a decision, optionally represented as a score, whether it is pristine or manipulated. The videos in the dataset are pre-processed such that the input to the network consists of a stack of an RGB frame containing the face crop detected in the video, followed by pre-computed optical flow fields from the 9 consecutive frames. The optical flow fields capture the motion between frames. The network architecture for DPNet is composed of three different components, a feature encoder, a prototype layer, and a class layer, which are described concisely below. For a more detailed explanation, the reader is referred to \cite{Trinh2021InterpretablePrototypes}.

The \emph{feature encoder} takes the input and extracts the features required for classification. This is done with HRNet \cite{WangSun2021}, a network with a large number of model parameters, which was designed to preserve high resolution images throughout the network. This is appropriate, as in deepfake detection, details are of the utmost importance. The feature encoder produces a latent representation of size $16\times 16 \times 256$ (height $\times$ width $\times$ depth) for each input sample.

The next component is the \emph{prototype layer} which calculates a similarity score between the latent representation of the input, and each of a predetermined number of prototype vectors for the two classes (pristine and manipulated). The prototype vectors are latent representations of crops from representative samples from the training set. These prototype vectors are learnable parameters of the network. We use 20 prototypes of size $1\times 1 \times 256$ for each class. For each spatial location in the latent feature vector, the similarity to the prototype is calculated. For each prototype, the highest similarity score is extracted. We use the same similarity function as in \cite{Trinh2021InterpretablePrototypes}. 

In the final \emph{class layer}, the maximum similarity score for each prototype is used as input to a fully connected classifier which outputs a score for the fragment being pristine, and being manipulated. 

\begin{figure*}[ht]
 \centering
 \includegraphics[width=\textwidth,keepaspectratio]{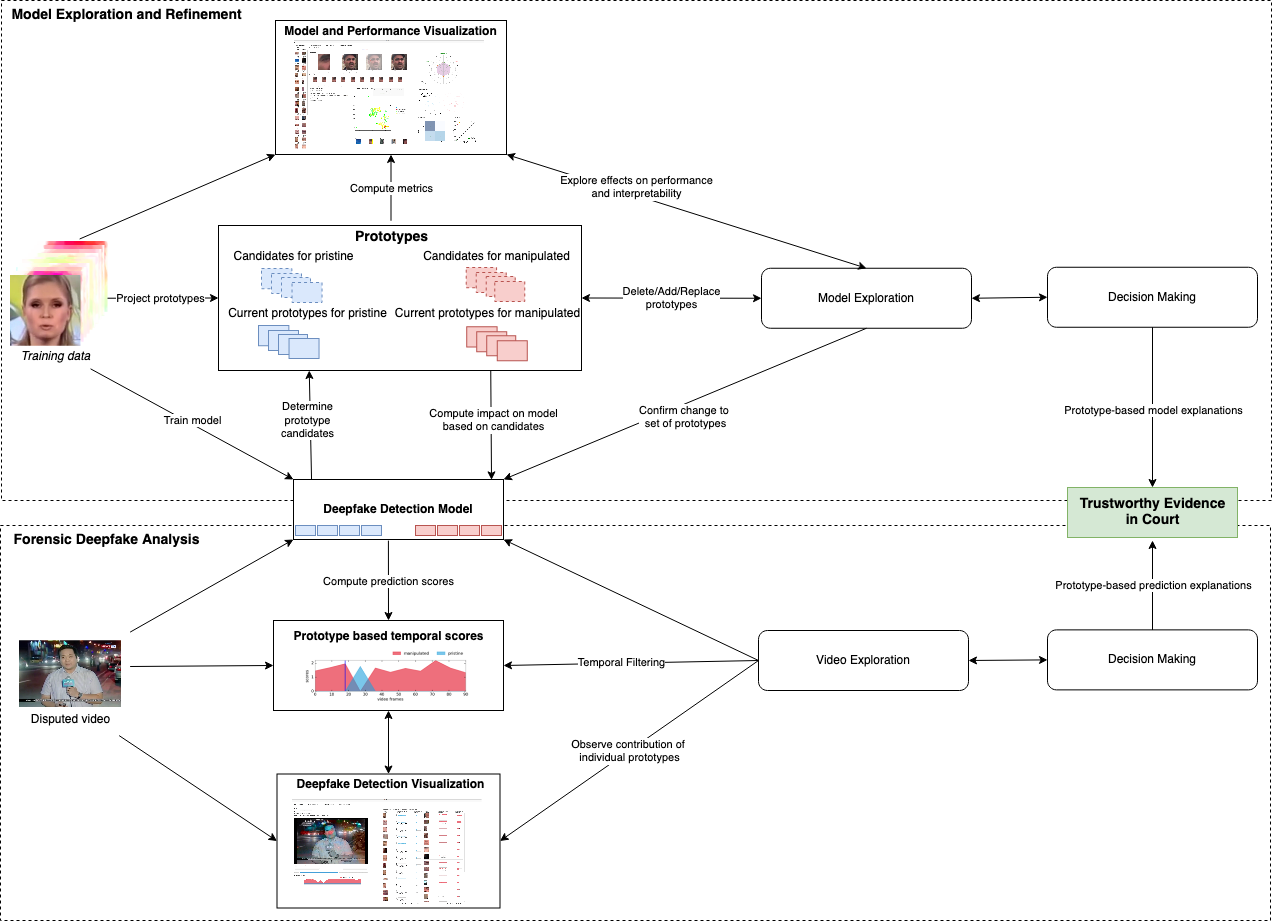}
 \caption{Diagram of our VA model for prototype-based interactive deepfake detection inspired by the knowledge generation model for visual analytics defined by Sacha \cite{sacha2014}. The diagram shows and models the two main tasks ProtoExplorer can aid the forensic expert in. Through model and video exploration the experts can make informed decisions at the model and video level, leading to trustworthy evidence in court.}
\label{fig:va_process_action_findings}
\end{figure*}

\subsection{Training the network}
We largely follow the training procedure carried out in \cite{Trinh2021InterpretablePrototypes}. They use a loss function consisting of four different components. A \textit{cross-entropy} loss term is used to penalize the misclassifications the network makes. A \textit{separation} loss term pushes the latent patches of the training images away from the prototypes that do not belong to the appropriate category. A \textit{cluster} loss term attempts to ensure that each training image has a latent patch that is close to at least one prototype of the appropriate class. A fourth component, the \textit{diversity} loss term, aims to prevent the overlap of prototype vectors from the same class. The different weights affect the relative importance of the different loss functions. 

Before training the network, the prototype vectors are randomly initialized vectors. During training they are optimized, but with no direct relation to actual video content. After a certain number of regular training steps (in our case, 5), a prototype projection step takes place. In this step, the training algorithm goes through all image sequences in the pre-processed training set and inputs them into the network, comparing the generated feature patches with the prototype vectors. For each prototype vector, the feature patch with the smallest distance to a prototype vector is selected, and those feature patches become the prototypes. This grounds the predictions of the model to the image sequences in the training set.

After each prototype projection step we optimize the weights in the last layer, similar to what is done in \cite{Chen2018ThisRecognition}. We freeze the parameters in all other layers and optimize a combination of the cross-entropy loss term and an $L_1$ loss term that encourages the weights of the connections between prototypes of the pristine class to the final manipulated score, and vice versa, to go to zero. 

\subsection{Efficient retraining after prototype refinement}
In the interface, the user will be able to delete and replace prototypes. After changing the model in this way, the network must be retrained. When the prototype vectors change, the weights in the final layer of the network must be optimized again. Backpropagating through the entire network is inefficient. We note that, if we save the intermediate outputs of the network during the initial training phase, we can use these to train only the last layer of the network. In case of a deletion, we use the saved maximum similarity scores for efficient training. In case of a replacement, we use the output of the feature encoder (because new similarity scores need to be calculated). Due to this change, retraining and testing the model can be done in real time.

\section{System Design}

The architecture of DPNet was conceived to automatically assign scores to the labels pristine and manipulated. The use of prototypes in DPNet makes it easier for human experts to understand how the decisions are made by the system. This process of understanding the model is completely post-hoc, i.e., it takes place after the model has been fully trained. The only way for the user to interact with the system is to change the weights of the loss terms and retrain the model. These weights steer the loss function of the model and consequently the prototypes of the model are changed. This indirect way of working with the prototypes in the model is difficult for a user as the system optimizes for representation learning and not for interpretability. We propose to make prototypes the prime object of interaction and make them the central part of our visual analytics process. 

\subsection{Design process}
The system is designed with and for video forensic experts. One of the co-authors of the paper is an expert witness in court on deepfake analysis, and based on his experience, informal conversations with other experts from the Netherlands Forensic Institute, and the literature review given in \cite{Westerlund_review}, the first version of ProtoExplorer was developed. This version was evaluated by five experts from the Netherlands Forensic Institute using think-aloud sessions while using the system (see the 'Evaluation' section for details), followed by interviews. Based on this, a second version of the system was built taking into account the comments. This  version was shown to the experts in subsequent in-depth interviews. This has led to the current version of the system as well as suggestions for future developments.

\subsection{Forensic deepfake analysis workflow}
The workflow for forensic experts is composed of three main tasks. The first task is \emph{video fragment selection}. Whether this task is present depends on the type of forensic question being posed. If the evidence is a large collection of videos, a selection has to be made of most likely candidates for manipulated fragments. For this automatic tools are most suited. In other cases, the expert receives a specific disputed fragment and should provide evidence whether this fragment is pristine or manipulated. Both cases lead to the second task which is \emph{forensic deepfake analysis} where an individual video fragment is analyzed for being manipulated. To provide trustworthy evidence in court this decision should be interpretable and the forensic expert should provide an understanding on how the model aided in making that decision. In our design this is represented by the \emph{model exploration and refinement} task. In this paper we focus on the last two tasks as this is where interactive solutions are most appropriate. Note, however, that a refined model can also give a better basis for informed decisions in the selection process. 

\subsection{Visual Analytics Process}

The overall diagram representing our process model is presented in Figure~\ref{fig:va_process_action_findings}. 

The aim of our interactive solution is to support the exploration and reasoning process of the forensic expert. To that end we rely mainly on the knowledge generation model defined by Sacha \cite{sacha2014}. This model shares the four main components (data, visualization, models, and knowledge) in the process model for VA defined by Keim \cite{Keim2008} and expands it by making human reasoning part of the VA process. In our case we model the tasks as intertwined decision making processes focusing respectively on the video and the model. Together they lead to informed decisions for trustworthy evidence in court. The two processes are coupled through the deepfake detection model, in particular DPNet. The whole model is centered around the prototypes in the detection model which on the one hand are the basis for exploring and refining model characteristics and on the other hand provide a decomposition of the prediction scores in the automatic analysis of a video fragment.  

From an interaction point of view the forensic deepfake analysis is relatively simple. It shows the prediction scores for the two classes for every frame in the video. The main action the forensic expert can perform is temporally zooming in on the part of the video that is of interest. By observing the contribution of individual prototypes to the overall prediction score, the expert can explain the decision made in terms of the prototypes contributing most. 

The model exploration and refinement task is more involved as it requires understanding the complex deep learning-based detection model. To reduce the cognitive load on the expert in understanding the model, we focus the exploration on the prototype layer in the model for two reasons. Firstly, this is the easiest layer to map to the characteristics experts use to manually analyze disputed videos and, secondly, prototypes are known to play an important role in cognitive processes \cite{Ware2004, Ware2008}. The interaction with the model revolves around the set of prototypes the system is using and a set of candidate prototypes the model provides. The exploration of the model consists of considering the set of current prototypes, deleting ones that are not deemed useful, or replacing them with a candidate prototype from the candidate set that might make the predictions more interpretable or less biased. However, changing the prototypes directly influences the performance of the detection model, so the visualization should emphasize the impact of the choices made. It is up to the expert to balance the quality of the detection versus the interpretability of the resulting model. 

By making the prototype layer the primary target of interaction, we create a methodology where the expert can directly manipulate the deepfake detection model, while keeping a large part of the model intact. In particular, we are not altering the large parameter set of the encoder. 

\begin{figure*}[ht]
\begin{tikzpicture}
\node[anchor=south west] (image) at (0,0) {
    \includegraphics[width=\textwidth]{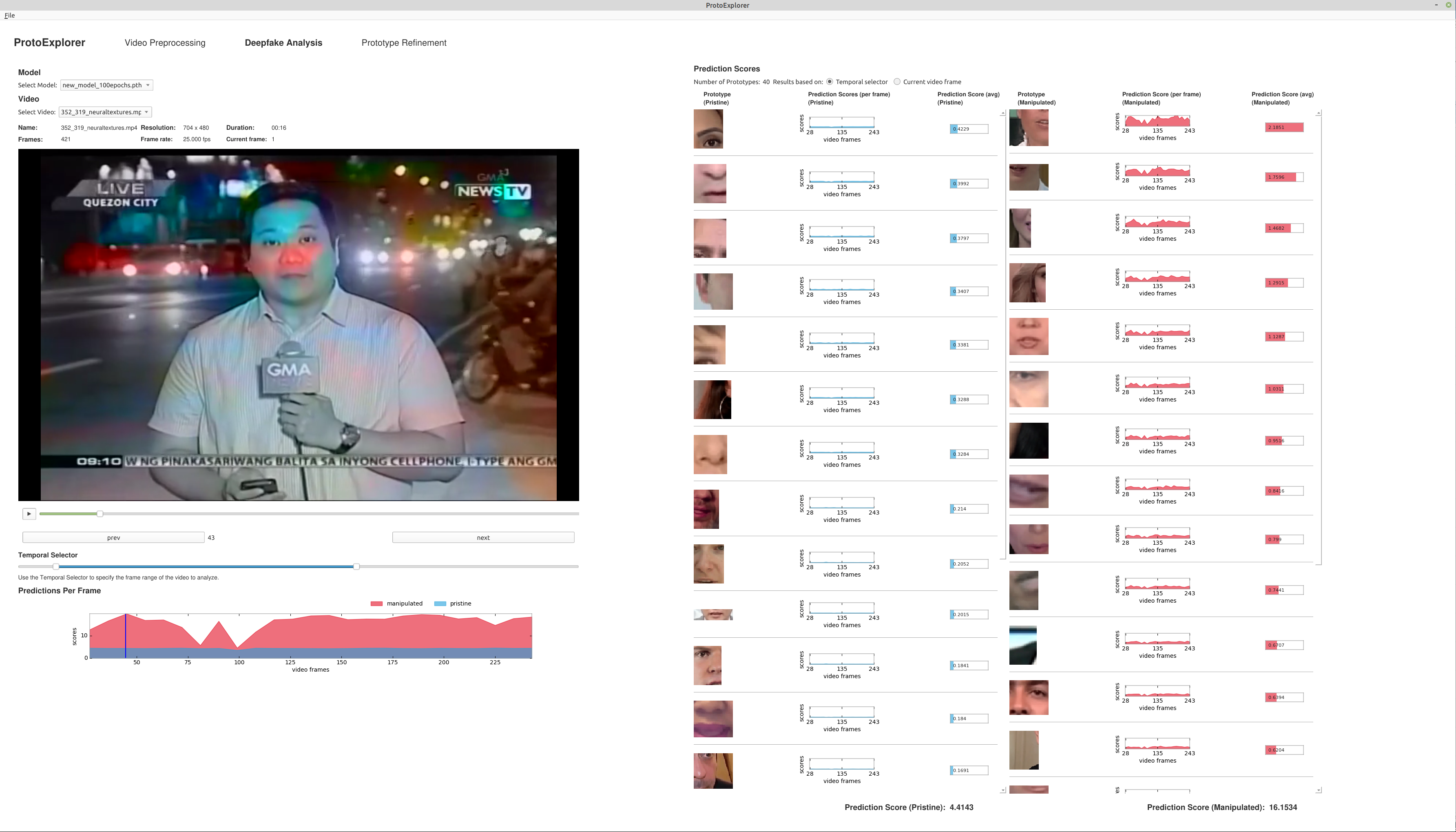}};
\node[circle,fill=yellow,minimum size=0.6cm] at (0.05,9.2){\tiny A};
\node[circle,fill=yellow,minimum size=0.6cm] at (0.05,8.6){\tiny B};
\node[circle,fill=yellow,minimum size=0.6cm] at (0.05,6.28){\tiny C};
\node[circle,fill=yellow,minimum size=0.6cm] at (0.05,3.3){\tiny D};
\node[circle,fill=yellow,minimum size=0.6cm] at (0.05,2.5){\tiny E};
\node[circle,fill=yellow,minimum size=0.6cm] at (8.0,9.2){\tiny F};
\node[circle,fill=yellow,minimum size=0.6cm] at (12.5,9.2){\tiny G};
\node[circle,fill=yellow,minimum size=0.6cm] at (8.65,0.0){\tiny F1};
\node[circle,fill=yellow,minimum size=0.6cm] at (10.2,0.0){\tiny F2};
\node[circle,fill=yellow,minimum size=0.6cm] at (11.7,0.0){\tiny F3};
\node[circle,fill=yellow,minimum size=0.6cm] at (12.5,0.0){\tiny G1};
\node[circle,fill=yellow,minimum size=0.6cm] at (14.0,0.0){\tiny G2};
\node[circle,fill=yellow,minimum size=0.6cm] at (15.5,0.0){\tiny G3};
;
\end{tikzpicture}
      \caption{
      The \emph{deepfake analysis screen} in ProtoExplorer.  (A) model selection; (B) input video selection and its metadata; (C) input video player; (D) video navigator including temporal selection; (E) visualization of the total prediction scores per frame for the selected temporal fragment of the video; (F)(G) visualization of the prototypes and prediction scores obtained by each prototype for both categories: \textcolor{pristine_color}{pristine} (columns F1, F2, and F3) and \textcolor{manipulated_color}{manipulated} (columns G1, G2, and G3);
 }
 \label{fig:deepfake-det-screen}
\end{figure*}
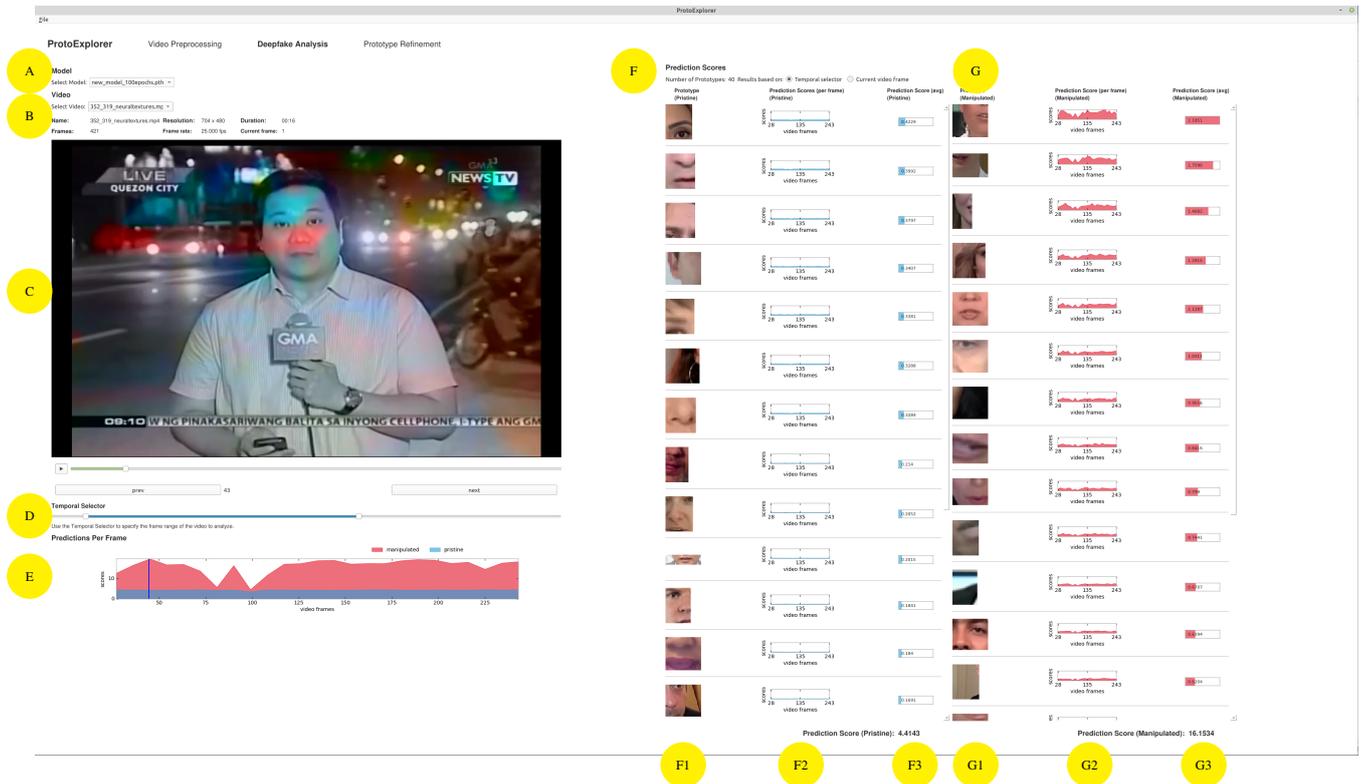

\section{System Overview}

We will now describe the visualizations developed to support forensic experts in performing their tasks. From there we will describe the two main interactive interfaces created to support their workflow.  

\subsection{Visualizations}

The layered architecture of the deepfake detection model provides a blueprint for the visualizations that are needed for the system. We consider here the main visualization components and do so from the perspective that prototypes form the core of our visual analytics model so these steer the different visualizations of the model.

\begin{figure}[h]
  \centering
    \subfloat{\includegraphics[width=.18\linewidth]{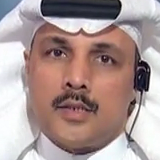}} \hspace{.1cm}    
    \subfloat{\includegraphics[width=.18\linewidth]{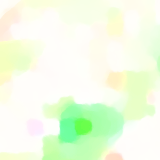}} \hspace{.1cm}       
    \subfloat{\includegraphics[width=.18\linewidth]{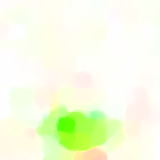}} \hspace{.1cm}        
    \subfloat{\includegraphics[width=.18\linewidth]{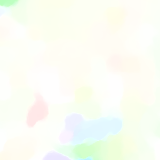}} \hspace{.1cm}        
    \subfloat{\includegraphics[width=.18\linewidth]{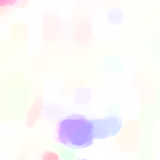}} \\    \vspace{-.2cm}
    \subfloat{\includegraphics[width=.18\linewidth]{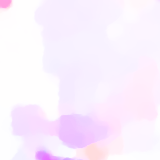}} \hspace{.1cm}  
    \subfloat{\includegraphics[width=.18\linewidth]{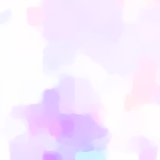}} \hspace{.1cm}  
    \subfloat{\includegraphics[width=.18\linewidth]{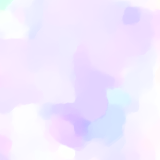}} \hspace{.1cm}  
    \subfloat{\includegraphics[width=.18\linewidth]{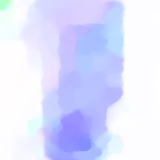}} \hspace{.1cm}  
    \subfloat{\includegraphics[width=.18\linewidth]{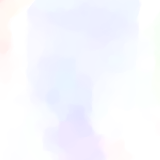}} 
  \caption{Sequence of the face crops generated in the pre-processing of an input video from the FaceForensics++ dataset. It includes ten frames: the first is the extracted face crop, and the rest were calculated by inputting the consecutive frames into the DeepFlow Optical Flow algorithm \cite{Weinzaepfel2013DeepFlow:Matching}.}
  \label{fig:optical_flow}
\end{figure}

\begin{itemize}
    \item \textbf{Prototypes:} The prototypes are the core of our visual analytics model and are spatio-temporal crops from the training videos. As in \cite{Trinh2021InterpretablePrototypes, Chen2018ThisRecognition}, we use the most highly-activated patch in the training video fragment as the visualization of the latent prototype vector. As they are taken from videos, the prototype vectors have a one-to-one correspondence to a specific video crop and can thus be visualized as such. We provide a number of different representations of the prototype. The main representation is the cropped area from the original video which can be viewed by its first frame or by an animated GIF. In addition, we show the prototype as a sequence of still frames as well as the corresponding flow field (see Figure~\ref{fig:optical_flow}) indicating which parts of the prototype exhibit movements which might also show inconsistencies of relevance. As context we provide the full original frame as well as the frame with an indication of where the crop is positioned. 
    \item \textbf{Embeddings:} The encoder part of the model maps individual spatio-temporal fragments into vectors which together form a high dimensional space. When the system optimizes this space based on the training samples it assures that similar samples are nearby and dissimilar ones are pushed away. This also holds for the prototype vectors. We can, however, not visualize this space directly. The UMAP projection methods \cite{UMAP} can map this space to two dimensions where it aims to preserve the relations between vectors as well as possible. We could do so for all training samples once and use that as a visualization, but as we are interested in the prototypes and the candidates we dynamically create the UMAP visualization for the current set. 
    \item \textbf{Similarity scores:} The overall contribution of each prototype is summarized in a single score which can simply be visualized as a number or any standard numeric visualization such as intensity. The similarity score is the global maximum of the similarity maps. These maps are abstract, but we use Prototypical Relevance Propagation \cite{Gautam2023PRP} to create activation maps that demonstrate what parts of the input video sequence contributed most to the highest similarity score. 
    Typically, the activation maps are generated by interpolating the similarity maps, which compare the latent features of the input video sequence to each of the prototypes. These activation maps are straightfoward to create but they give a global and spatially imprecise explanation, see for example \cite{Chen2018ThisRecognition}. In the final version of ProtoExplorer, we use Prototypical Relevance Propagation (PRP) \cite{Gautam2023PRP} to generate model explanations. PRP is an extension of Layer-Wise Propagation \cite{Bach2015LRP}, which is a post-hoc model explanation method that uses pre-defined computation rules to backpropagate the model output layer by layer, ultimately yielding a map distributing relevance over the input features. PRP aims to do the same, but it propagates the maximum similarity score for each prototype back to the input, effectively demonstrating what input features are responsible for this similarity score. Therefore, these activation maps for each prototype have a direct visual interpretation and let the expert see which part of the input sequence is contributing most to the partial classification corresponding to this prototype. 
    \item \textbf{Class layer:} In the class layer the contributions of individual prototypes are combined into a final score for the video fragment. For a specific video fragment the score for each prototype and the overall score is visualized as a time plot. The performance for the model as a whole, based on a test set independent from the training set, is visualized as a confusion matrix and an ROC curve. 
    \item \textbf{Model characteristics:} To characterize a specific instantiation of a model, i.e., a specific set of selected prototypes, we use a radar plot to summarize in one visualization the number of prototypes, the accuracy and area under the curve, as well as the different components of the loss function.  To make for a uniform presentation, we normalize all of these to the initial model the expert started with. It is important to note that an important characteristic of the model is the distribution of the different prototypes over the frame as every prototype is only a crop of the input sequence. Thus, it only triggers a restricted part of the activation map which can easily lead to a bias in the classification to certain regions of the frame. To that end we developed the landmark density map which shows how the prototypes are distributed in terms of relevant landmarks. Using this visualization we noticed that the standard model, even by tuning the weights corresponding to the individual loss terms and increasing the number of prototypes, does not yield an even distribution over the frame. This is another indication that an interactive solution is essential.
\end{itemize}

Having defined the core visualizations it is time to connect them to the proposed visual analytics process. This leads to two main interactive visual interfaces, corresponding to the forensic deepfake analysis task and the model exploration and refinement task.  


\subsection{Forensic deepfake analysis interface}
The deepfake analysis screen allows the user to conduct deepfake detection using interpretable prototype-based models. The screen is depicted in Figure~\ref{fig:deepfake-det-screen}.

The user starts by selecting the model from a list of pre-trained DPNet models (A) and a video from the list of pre-processed videos in the video panel (B) presenting some basic information on the video.  
The video player (C) allows the user to play, pause, scroll, and step frame by frame through the video via the playback timeline. With the temporal selector (D) the user can define the temporal fragment of the video to be analyzed. 

The predictions per frame (E) visualization shows the predictions for each of the two categories: pristine and manipulated. The total prediction scores are shown on the vertical axis, and frames of the video are on the horizontal axis. Initially, the horizontal axis displays all video frames.  
The prediction score panels (F,G) visualize the individual contributions of the prototypes (F1,G1) to the overall predictions. When the single frame mode is selected a visualization similar to the prototype viewer in the prototype refinement screen (see Figure~\ref{fig:deepfake-det-screen}) is given. For a temporal fragment, the contribution of each prototype to the total predictions per frame is given in (F2,G2) where the contribution to the final score is given in (F3,G3).  When the user adjusts the temporal video range in the temporal selector, all prediction charts are automatically updated.

Using this interface the expert can drill down to a specific fragment of the video that, according to the system, has the most prominent signs of being non-pristine material. The system explains itself in terms of the prototypes and how they contribute to the final score. By looking at the prototypes that contribute most, and possibly switching to the frame visualization mode where more details of the prototype are given, the expert can validate whether the prototypes indeed give a good explanation. Next to that the expert can also verify that the pristine part of the video is also recognized in the correct way. If the prototypes do not provide a convincing explanation because, e.g., no clear artifacts are visible in the manipulated class or if the prototypes are not diverse enough, the expert can decide to move to the model exploration and refinement interface to find a better set of prototypes.  

\subsection{Model Exploration and Refinement interface}
In the model exploration and refinement screen, the user can import a pre-trained initial model and conduct a series of operations to refine its prototypes. This screen consists of the following panels, shown in Figure~\ref{fig:refinement_screen}:
\begin{figure*}[ht]
\begin{tikzpicture}
\node[anchor=south west] (image) at (0,0) {
    \includegraphics[width=\textwidth]{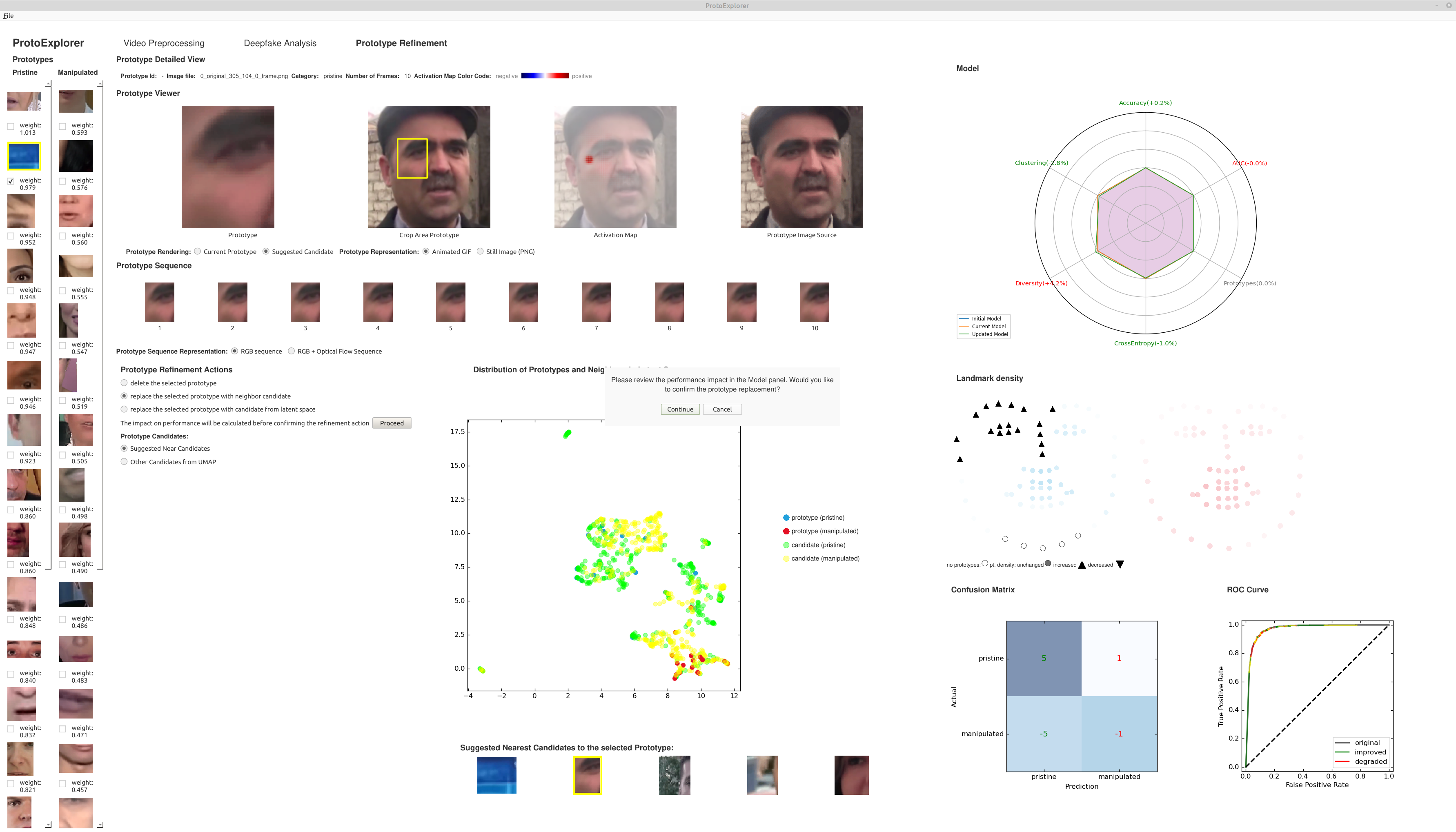}};
\node[circle,fill=yellow,minimum size=0.6cm] at (0.85 - 0.07,10.0 - 0.37){\tiny A};
\node[circle,fill=yellow,minimum size=0.6cm] at (2.0 - 0.07,10.0 - 0.37){\tiny B};
\node[circle,fill=yellow,minimum size=0.6cm] at (2.0 - 0.07,8.9 - 0.37){\tiny B1};
\node[circle,fill=yellow,minimum size=0.6cm] at (2.0 - 0.07,7.0 - 0.52){\tiny B2};
\node[circle,fill=yellow,minimum size=0.6cm] at (2.0,5.5){\tiny C};
\node[circle,fill=yellow,minimum size=0.6cm] at (2.0,4.5){\tiny C1};
\node[circle,fill=yellow,minimum size=0.6cm] at (5.5,5.0 + 0.3){\tiny C2};
\node[circle,fill=yellow,minimum size=0.6cm] at (5.5,2. - 0.6){\tiny C3};
\node[circle,fill=yellow,minimum size=0.6cm] at (11.4 - 0.2,10.0 - 0.37){\tiny D};
\node[circle,fill=yellow,minimum size=0.6cm] at (11.4 - 0.2,9.5 - 0.37){\tiny D1};
\node[circle,fill=yellow,minimum size=0.6cm] at (11.4 - 0.2,5.8 - 0.25){\tiny D2};
\node[circle,fill=yellow,minimum size=0.6cm] at (11.4 - 0.2,3.0){\tiny D3};
\node[circle,fill=yellow,minimum size=0.6cm] at (15.0 - 0.5,3.0){\tiny D4};
;
\end{tikzpicture}
  \caption{
 ProtoExplorer, a prototype based Visual Analytics system for analyzing deepfake videos, showing the \emph{model exploration and refinement screen}.  The list of prototypes (A) can be viewed in detail (B) and revised through deletion, replacement, or addition of prototypes (C), the impact of which can be observed in the performance panel (D). 
  }
  \label{fig:refinement_screen}
\end{figure*}
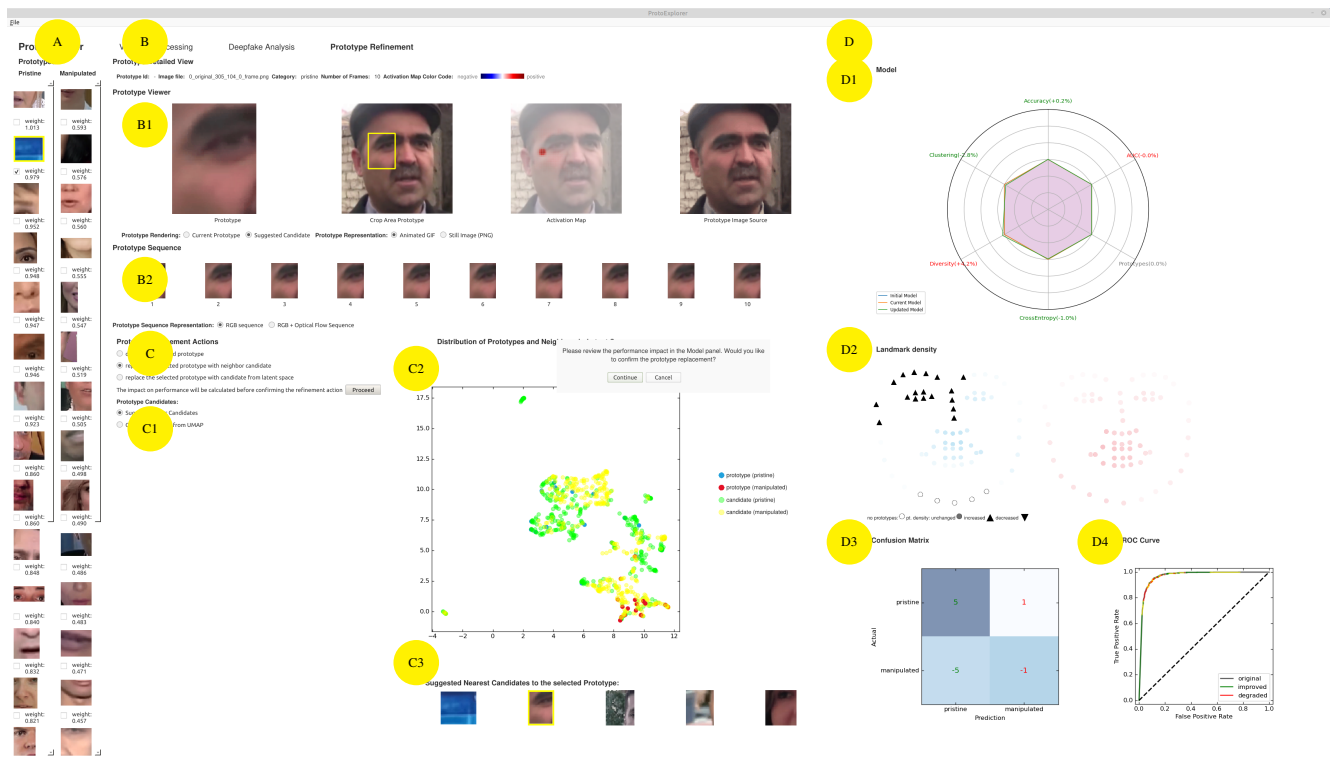

The prototypes panel (A) contains the prototypes of the model for the pristine and manipulated categories. These are displayed in two vertical scrollable lists so they can be viewed properly, even when the number of prototypes differs between models. Underneath each prototype, the weight associated with it in the final linear layer of DPNet is shown, and the prototypes are sorted according to their weights. The user can then select one prototype from the list to inspect in the prototype detail view (B), or several of them when aiming to delete multiple prototypes at once.
 
The prototype detail view offers different representations of the selected prototype. The prototype viewer (B1) consists of four display panels that visualize (from left to right): the \textit{selected prototype}, the dataset sample that was the \textit{source of the prototype} including its \textit{cropping area}, the \textit{PRP map} that determined the cropping area, and the same but unmodified \textit{dataset sample}. The user can choose to view the prototypes as animated GIFs or as still images. The prototype Sequence viewer (B2) consists of the ten consecutive images that were used to generate the prototype, either showing all individual frames of the prototype, or the first frame and the subsequent flow fields.

The prototype refinement panel (C) allows the user to delete or replace prototypes (C1). 
The user can delete one or more prototypes at once allowing for a faster refinement when working with models having a high number of prototypes. When reducing the number of prototypes is not desired, replacing a prototype with an alternative might help increase the interpretability of the predictions. There are two options for replacing a selected prototype. Candidates near to the selected prototype are displayed and can be selected in an image gallery (C3). Alternatively, a choice can be made from a larger set of prototype candidates in an interactive UMAP \cite{UMAP} visualization (C2) alongside the selected prototype. When a prototype candidate is selected, it is displayed in the prototype viewer. 
As the UMAP gives a convenient overview of the relative positioning of prototypes and candidates, the expert can choose candidates in regions of the space that are under-represented. 

The performance panel (D) gives an overview of the performance of the current model and prototypes in a radar mplot (D1), landmark density plot (D2), confusion matrix (D3), and ROC curve (D4). This provides the expert with information on where improvements are necessary in terms of a better classification performance (in the confusion matrix), a better balance between false and true positives (in the ROC curve), more diverse, clustered, or separated distribution of prototypes (the radar plot) or an improved distribution over the landmarks in the image (the landmark density plot). 

When the user changes the set of prototypes it impacts the performance of the model as well as several other characteristics of the model. Before the user confirms the change, the system computes the characteristics of the model by evaluating the model on a test set independent from the training set used for DPNet. This takes a few seconds. The result of the evaluation is shown in the performance panel. As the task of the expert is to assess the impact of the change, the visualizations switch to a mode in which differences are highlighted. The radar plot shows three curves namely the initial model (where all characteristics are displayed as 100\%), the current model, and the new model. The labels provide the relative changes of the new model with respect to the current model. The landmark density plot gives an indication of which landmarks are affected and whether they go up or down. In the confusion matrix the actual change is indicated. To make it clear whether these are improvements or degradations, the color of the text is made green or red respectively. The ROC curve simply shows the two curves at the same time as this is a familiar way of viewing the improvement of a model. 
 
\section{System Implementation}
\label{sec:system_implementation}
In this section we describe the technologies and methods used to implement the system, and to train the prototype-based deep learning models included in the system. The system consists of a standalone desktop application developed in Python. The framework used for the Graphical User Interface is Qt in its python binding.

The models trained for ProtoExplorer are based on the Dynamic Prototypes Net (DPNet) \cite{Trinh2021InterpretablePrototypes}. As mentioned in the background section, DPNet's neural network architecture includes HRNet \cite{WangSun2021} as a feature encoder, which was pre-trained on ImageNet \cite{ImageNet}. To train these models, we used weights $\lambda_{clus}=0.2$, $\lambda_{sep}=0$ 
, $\lambda_{div}=0.1$, $\lambda_{L1} = 0.001$ 
and randomly initialized prototypes vectors as specified in \cite{Trinh2021InterpretablePrototypes}. We used 20 prototypes for the pristine class and 20 prototypes for the manipulated class.

The network was trained on the HQ (c=23) variant of the Face Forensics++ \cite{Rossler2019FaceForensics++} dataset, broadly used in deepfake detection. Although Face Forensics++, in its latest revision, includes five different deepfake generation techniques for training the models we used the four that were included in its initial release and used them to train DPNet \cite{Trinh2021InterpretablePrototypes}, namely Deepfakes \cite{DeepfakesGithub}, Neural Textures \cite{Thies2019DeferredTextures}, FaceSwap \cite{FaceswapGithub}, and Face2Face \cite{Thies2020Face2Face:Videos}.

Before the network was trained, the training set from FaceForensics++ needed to be pre-processed. Once the frames of the input video were extracted, MTCNN \cite{Zhang2016JointNetworks} was used for face detection, storing bounding box coordinates of the detected faces. We sampled a frame every \textit{k=10} frames. Then, we used the sampled frame and its consecutive $k-1$ frames to generate $k-1$ flow frames (Figure~\ref{fig:optical_flow}) with the optical flow algorithm DeepFlow \cite{Weinzaepfel2013DeepFlow:Matching}. Animated GIF image files were then exported for each of the time sequences to enable their visualization in the model exploration and refinement and deepfake analysis screens. The sampling of frames from the videos of the dataset was conducted following the instructions in \cite{Trinh2021InterpretablePrototypes}. 

\section{Evaluation}
\label{sec:evaluation}
To validate our system we involved five anonymous video forensic experts. All the experts have more than nine years of experience at the Netherlands Forensic Institute. They all have experience with doing deepfake detection in the forensic setting and have knowledge of deep learning models. In their current workflow they do use different automatic deepfake detection tools, but rely heavily on visual inspection of the disputed video to make their judgements. It should be noted here that it would be difficult to take their current systems as baselines for a direct comparison to ProtoExplorer. There would be too many dimensions which are different starting from the training data used for the different systems, the underlying deep learning model, and the different ways of interacting with the system. Furthermore, the systems in use have no visual analytics component so cannot interact with the model directly. 

The evaluation was done in two rounds. The first round was based on sessions where the experts interacted with the system and in-depth interviews while the second round was based on interviews only.  The first round was the first time the experts experienced the graphical user interface of ProtoExplorer. We will now elaborate on the setup and results for those rounds and how they have led to improvements to the intermediate versions of the system and finally how they lead to ideas for further improvements.

\subsection{First evaluation round}
The evaluation took place in individual sessions with a duration of 90 to 110 minutes. In the first 10 minutes, prototype-based models were introduced to the forensic expert. For the next 10 to 15 minutes a demo was presented showing how to explore the prototypes of the model, refine the model by replacing and deleting prototypes, and use the model to conduct deepfake detection. The experts were then invited to perform a series of four interactive tasks for 30 to 40 minutes. As protocol we used the `Thinking aloud' \cite{JakobNielsen1993UsabilityEngineering} method for usability testing, asking the users to verbally express their thoughts while they interact with the system. The audio containing their spoken thought process was recorded. 

The tasks given to the experts were as follows:
\begin{enumerate}
    \item \textbf{Understand the model}: Explore the model in the \emph{model exploration and refinement interface} as provided by the system. This includes the inspection of its prototypes in the different visualizations and considering the performance metrics as provided in the confusion matrix and ROC curve.
    \item \textbf{Deepfake analysis using the current model} Using the initial model, analyze the disputed video using the \emph{forensic deepfake analysis interface}. This can be done using the different modes to calculate predictions, and by temporally exploring the video. The user did not know whether the video was pristine or manipulated.
    \item \textbf{Prototype Refinement}: Use the delete / replace refinement operations in the \emph{model exploration and refinement interface} to steer the model into using prototypes that give more interpretable predictions, while taking into consideration the impact of these refinement operations on the performance of the model. 
    \item \textbf{Compare initial and refined model}: Analyze the disputed video again using the \emph{forensic deepfake analysis interface} and compare the results of the initial and the refined model.
\end{enumerate}

After the session we interviewed the expert for 15 to 20 minutes, asking questions about their impression of the system and how to improve different aspects of the system. The audio recordings of the interactive sessions and the interviews were transcribed and manually categorized into six different categories namely `Prototype Exploration', `Prototype Refinement', `Model information and metrics', `Deepfake Analysis', `Interpretability and Forensics' and `General Remarks'. 

\subsection{Prototype Exploration}
In the first task of the evaluation when the expert users were asked to inspect the prototypes of the given model, they made extensive use of all the options available in the prototype detailed view, switching back and forth between visualizing the image sequence and specific frames of the prototypes. 
Their interaction with the system was fluent. The system enabled them to intuitively visualize and explore the prototypes of the model. The feedback was positive, with a few suggestions for future work such as allowing the user to hide panels to maximize screen space, re-positioning some selectors, and adding interpolation-free image scaling options.

\begin{figure*}[ht]
  \centering
    \includegraphics[width=\textwidth]{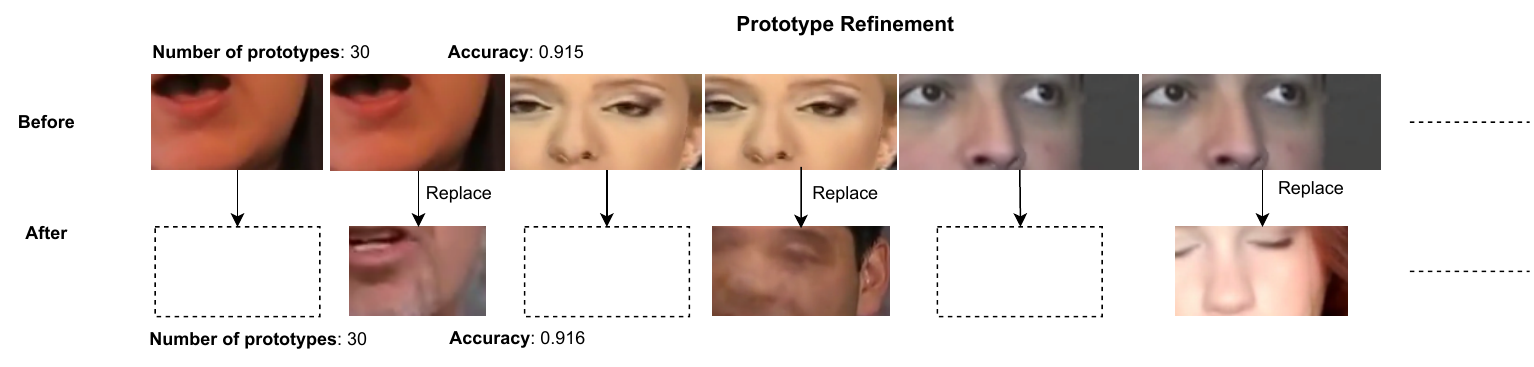}
      \caption{
    Information about the prototype refinement, including the impact on the accuracy, conducted in the evaluation by one of the experts. The top row shows the original prototypes which exhibit nearly identical prototypes. The expert selected three new prototypes to replace the similar ones and thus increased diversity while getting (slightly) better accuracy. 
 }
 \label{fig:proto-ref-case-study}
\end{figure*}

\subsubsection{Deepfake Analysis}
The expert users appreciated the functionalities offered by our system for deepfake analysis. To inspect the video under investigation, they used both prediction modes: based on the temporal selector and based on the current video frame selected in the video player. When asked, they found both prediction modes useful.
When analyzing the results based on the temporal selector, they highlighted the benefits of the system allowing them to interactively select and adjust the frame range for analysis and calculate an average of the predictions. One of the experts mentioned that this functionality was not available in other software for deepfake detection that they were familiar with. 
The video forensic experts highlighted the usefulness of being able to display the predictions scores for both categories at the same time, which made it possible for them to easily compare the total scores and the prototypes with a higher contribution to the scores.
When they felt the need to take a closer look at a specific frame sequence of the disputed video, the expert users appreciated that the system provided predictions for the video frame selected in the video player.  
When using this option, the forensic experts were able to visually analyze the regions of the input crops that are compared to the prototypes. Suggestions were for a large part on basic elements like the video player and its relation to the temporal visualization. 


\subsubsection{Interpretability and forensics}

An important consideration for the experts is that to defend the decision in court they have to make clear that they looked at all aspects of the face like eyebrows, mouth, teeth, and hairline to substantiate their decision. The prototypes automatically derived by the system did not cover all of those regions and thus it is important to add additional prototypes to increase the diversity. As indicated earlier they are, however, not always available for selection. 
The interpretation of the comparison between the input and prototypes of the network that led to the scores based on similarity was a recurrent topic of discussion in the evaluation of the system. The thorough evaluation with the video forensic experts indicated that the promise of prototype-based methods of providing human understandable predictions based on visual similarity could not reach its full potential when using spatio-temporal prototypes. The activation maps generated when applying the model to a disputed video determining the patches of the input face crops that are compared to the prototypes, do not always compare the same facial regions. One of the reasons for this could be that the model is not only comparing visual but also temporal information captured in the optical flow fields. This has a negative impact on the explainability of the predictions. Although the flow fields are visualized it is difficult to grasp this aspect.


\subsubsection{Prototype refinement}
Another topic of discussion during the evaluation of our system was the criteria that should be followed for deciding whether a prototype should be replaced or deleted. The duplication of prototypes, or the inclusion of prototypes that are almost identical, was the one that generated consensus among experts. The experts either replaced or deleted duplicate prototypes. When replacing was the choice, see Figure~\ref{fig:proto-ref-case-study}, the forensic expert evaluated the prototype candidates suggested by the system. 
When analyzing the candidates for a prototype belonging to the manipulated category, an approach mentioned by the forensic experts was to look for prototypes that contain representative visual artifacts of deepfakes.
The other criterion that was mentioned was to consider prototype candidates that  provide a more diverse prototype selection. Several experts noted that the prototype selection does not necessarily cover all facial regions.
This was highlighted multiple times during the evaluation as a factor that has a negative impact on the interpretability of the predictions. Clearly, some experts, when evaluating candidates for replacing a prototype, looked for prototypes that contained facial regions that were not yet represented in the selection. 

Another aspect that was mentioned was that the prototypes from one category cover a different facial region (e.g., nose) than the other category (e.g., chin) 
This was considered by the experts to have a negative impact on interpretability. They suggested adding functionality that makes it possible to search for prototype candidates based on the facial region, as well as other criteria such as the angle of the face, or skin tone.

During the refinement of the model, calculating the impact of the changes in the prototype selection was a key design requirement. This provides relevant information to the expert user before committing changes in the model. The feedback received during the evaluation reaffirmed the utility of this functionality, but the experts suggested reducing the number of waiting intervals.
In cases when the expert user is certain about the need for a refinement operation, it would be helpful to be able to immediately make the change, without any calculation. 
This could create a more fluent refinement process for the user. The use of entire new sets of similar prototypes (that maintain the performance of the model) rather than focusing on replacing individual prototypes was also suggested as an option which could minimize the waiting time needed to refine the model.

When searching for prototypes, the experts were often also talking in terms of semantic attributes of the input video sequences like having prototypes corresponding to a certain skin color or angle of the face like a side view instead of a frontal view. 

The candidates that were suggested by the system, namely the neighbors of the current prototype, were difficult to interpret (and often quite similar) and it was difficult to make a choice for one of those. A suggestion made by one expert was to start without any prototype and start adding prototypes from specific artefacts found one-by-one. Although this seems attractive, the setup of a prototype based system like DPNet requires a set of initial prototypes for the model. 

\subsubsection{Model information and metrics}

Apart from some general comments on the number of decimals for measures or whether they were given in percentages or not the main discussion elements were focused on two parts. The first was the indication of change in the confusion matrix. It was not always clear that these were relatively compared to the current model or the actual performance of the candidate model. Suggestion was to use colors to indicate whether the change was considered desirable or not. In general it should be more clear that the model has changed. The second was the use of the word similarity for the contribution of a prototype to the final result and whether high was better or worse. This led to some confusion. 

\subsection{General remarks}

In the general remarks two recurring topics were mentioned. Firstly, there are plenty of useful visualizations but for many users it is sufficient to simply have an indication that the video is pristine or manipulated. This is similar to the previously mentioned use of automatic methods for selecting videos. However, that is not the scope of the current paper. What is possible is making certain visualizations optional. The second one has to do with the efficiency of testing the model given a set of prototypes. At the time of the interviews, this took 60-120 seconds. Suggestions for improvement were to perform multiple prototype refinement steps (like deleting multiple prototypes) before calculating the impact on performance. 

\subsection{System improvements after first evaluation round}

The first evaluation round led to a number of improvements to the system. Some of the changes were simple like making the layout more consistent, grouping elements in a more logical way, and adding some elements to the video player to allow stepping through, frame-by-frame. Other changes were more involved. 

One of the main changes we implemented was a speedup of the training and testing procedure which takes place after prototype refinement. In the initial version of ProtoExplorer, the weights in the last layer were not optimized after prototype refinement. Simply testing the model on the testing data took minutes, and training was not considered. In the most recent version, we saved the intermediate outputs of the first model to be able to efficiently retrain and test refined models, as described in the Background Section. This reduces the combined retraining and testing time to less than a second for prototype deletion and a few seconds for prototype replacement. 

To improve the selection of candidate prototypes we added the UMAP visualization to give an overview of the different candidates and their distribution in the embedding layer. In this way the investigator can more easily spot areas that are not represented or find close elements having different labels to allow for more subtle distinctions (see (F4) in Figure~\ref{fig:refinement_screen}).

A second focus of changes was the model performance and metrics panel. Here we made the suggested changes to the confusion matrix, adding green/red colors when used in comparison mode. We also added the radar plot to have an instant view of the changes in the model characteristics as a whole (replacing a text based indication of multiple characteristics). One of the most important changes was developing the landmark density visualization as the distribution of prototypes over different regions of the face was one of the major discussion points in the evaluation. 

The experts found the interpolated activation maps as typically used in prototypical networks lacking in clarity and spatial precision. Based on this, we replaced these by maps generated using Prototypical Relevance Propagation, as described in the 'System Implementation' section.

\subsection{Second evaluation round}

The second round of interviews confirmed the usefulness of (most of) the changes, identified some minor issues, and led to a number of new directions to pursue. Two interrelated elements were coming back in the different interviews, namely expertise and workflow. The current system is mostly geared towards forensic experts, and it would be good to make a different profile for police officers. A more explicit visualization of the workflow in using the system would help for both profiles. As one expert stated, what the experts need to do in the end is simply being able to state which frames in the video are suspicious and in what part of the frame the manipulation is visible. The steps in the workflow should lead the expert to that decision. 

Although the experts indicated in the first round that more elaborate search methods for prototypes are needed they found it difficult to understand what was depicted in the UMAP visualization. More guidance to help them pick candidates from this visualization is needed. 

Additionally, there were some desired functionalities which define interesting avenues for further research. These all revolve around the selection of prototypes. Notable improvements include being able to select candidate prototypes based on semantic characteristics such as putting prototypes at the mouth or eyelashes at the top of the candidate list, and providing a ranking of prototypes based on their impact on the different performance metrics. 

\subsection{Discussion and future work}
Although ProtoExplorer was developed for forensic experts, it also yields good opportunities for further development of prototype-based automatic deepfake detection methods. Building models using DPNet and visualizing them in ProtoExplorer revealed that although classification performance is quite good the prototypes are not always distributed over the whole face. To study this problem we should aim to vary the weights corresponding to the different loss terms for the automatic system and evaluate the result using our landmark density visualization to see the distribution on the face, our radar plot to see the impact on loss functions, and our UMAP visualization to see the distribution in latent space. As mentioned in the 'System Implementation' section, the dataset we used for training the models was FaceForensics++. This is the dataset used in  \cite{Trinh2021InterpretablePrototypes} to benchmark the performance of the DPNet network that we used. A recent study \cite{FairnessDeepfakeDatasets2021} concluded the existence of a representation bias in this dataset in terms of race and gender, a phenomenon we also saw in using the system. To improve on this we would have to automatically label the training data for those characteristics. From there we could see whether the above explorations could contribute to less bias or that additional loss terms should be considered.  

As mentioned in the `System Implementation' section, the models used by our systems were trained following the DPNet network introduced in \cite{Trinh2021InterpretablePrototypes}. This network, rooted in ProtoPNet \cite{Chen2018ThisRecognition}, used spatio-temporal prototypes instead of ones based exclusively on still images. The addition of this temporal information is the most significant contribution of this method, aiming to capture temporal inconsistencies present in deepfake videos. Based on the performance comparison in \cite{Trinh2021InterpretablePrototypes}, DPNet outperforms ProtoPNet in deepfake detection, initially validating the intuition that temporal information would be beneficial for this task. However, this decision had pitfalls that were only revealed when explored in an interactive setting. 

The evaluation with the experts indeed revealed the gap between the system prototypes that are good for classification and the prototypical examples experts use. In this paper we took the automatic system as the starting point and let experts refine the prototypes. An interesting alternative would be to let experts build up a rich vocabulary of prototypical situations with good spatio-temporal examples and see how the system can be adapted to take those as the starting point. The expert could then use ProtoExplorer to browse these prototypes  whenever they are presented with new disputed material. 

\section{Conclusion}
In this paper, we proposed a Visual Analytics process model for prototype learning inspired by \cite{sacha2014} composed of two intertwined decision-making loops. The model is specifically designed for deepfake analysis but gives a blueprint for any scenario in which there is a balance needed between optimal performance and interpretability using a prototype-based deep neural network. Based on the model we presented ProtoExplorer, a VA system allowing for the exploration of prototypes and the refinement of prototype-based models for the forensic detection of deepfake videos. Our system facilitates the exploration, deletion, and replacement of complex dynamic prototypes that combine spatial and temporal information.
We conducted an evaluation with five video forensic experts in which we assessed the contributions of ProtoExplorer. The evaluation reaffirmed that ProtoExplorer is a significant step in the right direction of closing the gap between the characteristics experts use in analyzing disputed videos and automatic methods. The evaluation gave insights into the complex challenges remaining in achieving highly interpretable, prototype-based predictions in the analysis of video sequences and yielded a number of promising new research directions for deepfake analysis as well as for more generic visual analytics using prototype-based systems. 

\begin{acks}
The authors wish to thank the anonymous video forensic experts from the Netherlands Forensic Institute who participated in the evaluation of ProtoExplorer.
\end{acks}

\bibliographystyle{SageV}
\bibliography{ms.bib}

\end{document}